\documentclass[conference]{IEEEtran}
\IEEEoverridecommandlockouts
\usepackage{cite}
\usepackage{amsmath,amssymb,amsfonts}
\usepackage{algorithmic}
\usepackage{graphicx}
\usepackage{textcomp}
\usepackage{xcolor}
\usepackage{algorithm}
\usepackage{algorithmic}
\usepackage{booktabs}
\usepackage{fancyhdr}

\usepackage[colorlinks=true, linkcolor=blue, citecolor=blue, urlcolor=blue]{hyperref}

\def\BibTeX{{\rm B\kern-.05em{\sc i\kern-.025em b}\kern-.08em
    T\kern-.1667em\lower.7ex\hbox{E}\kern-.125emX}}

\makeatletter
\def\BibTeX{{\rm B\kern-.05em{\sc i\kern-.025em b}\kern-.08em
    T\kern-.1667em\lower.7ex\hbox{E}\kern-.125emX}}
\makeatletter
 \let\old@ps@headings\ps@headings
 \let\old@ps@IEEEtitlepagestyle\ps@IEEEtitlepagestyle
 \def\confheader#1{%

 \def\ps@IEEEtitlepagestyle{%
 \old@ps@IEEEtitlepagestyle%
 \def\@oddhead{\strut\hfill#1\hfill\strut}%
 \def\@evenhead{\strut\hfill#1\hfill\strut}%
 }%
 \ps@headings%
 }
 \makeatother

\fancypagestyle{firstpage}{
  \fancyhf{} % Clear all header and footer fields
  \fancyhead[L]{}
  
}

 \usepackage[pscoord]{eso-pic}
\newcommand{\placetextbox}[3]{
 \setbox0=\hbox{#3}
 \AddToShipoutPictureFG*{ \put(\LenToUnit{#1\paperwidth},\LenToUnit{#2\paperheight}){\vtop{{\null}\makebox[0pt][c]{#3}}}
 }
 }
 \placetextbox{.23}{0.055}{\small{}}
 
\begin{document}

\title{Integrating Machine Learning Ensembles and Large Language Models for Heart Disease Prediction Using Voting Fusion\\

\author{
Md. Tahsin Amin$^{1,\dagger}$
Tanim Ahmmod$^{2,\dagger}$,,
Zannatul Ferdus$^{3}$,\\
Talukder Naemul Hasan Naem$^{2}$,\\
Ehsanul Ferdous$^{2,*}$,
Arpita Bhattacharjee$^{4,*}$\\
Ishmam Ahmed Solaiman$^{5}$, Nahiyan Bin Noor$^{6}$\\
$^{1}$Faculty of Computer Science \& Engineering,
Patuakhali Science \& Technology University, Bangladesh\\
$^{2}$Dept. of Electrical \& Electronic Engineering,
Rajshahi University of Engineering \& Technology, Bangladesh\\
$^{3}$ Dept. of Industrial \& Production Engineering,
Ahsanullah University of Science \& Technology, Bangladesh\\
$^{4}$ Dept. of Computer Science \& Engineering,
National Institute of Technology, Silchar, India\\
$^{5}$ University of Arkansas at Little Rock, Little Rock, Arkansas, USA\\
$^{6}$ Institute for Digital Health \& Innovation, University of Arkansas for Medical Sciences, Arkansas, USA\\
Email: tahsin16@cse.pstu.ac.bd,
tanim.eee.ruet@gmail.com,
zferdus508@gmail.com,\\
naemruet@gmail.com,
md.ehsanulferdous@gmail.com,
arpitabhattacharjee1934@gmail.com
}

\thanks{$^{\dagger}$These authors contributed equally to this work.}
\thanks{$^{*}$These authors also contributed equally to this work.}
}

\maketitle
\thispagestyle{firstpage} %
\begin{abstract}
Cardiovascular disease is the primary cause of death globally, necessitating early identification, precise risk classification, and dependable decision-support technologies. The advent of large language models (LLMs) provides new zero-shot and few-shot reasoning capabilities, even though machine learning (ML) algorithms, especially ensemble approaches like Random Forest, XGBoost, LightGBM, and CatBoost, are excellent at modeling complex, non-linear patient data and routinely beat logistic regression. This research predicts cardiovascular disease using a merged dataset of 1,190 patient records, comparing traditional machine learning models (95.78\% accuracy, ROC-AUC 0.96) with open-source large language models via OpenRouter APIs. Finally, a hybrid fusion of the ML ensemble and LLM reasoning under Gemini 2.5 Flash achieved the best results (96.62 \% accuracy, 0.97 AUC), showing that LLMs (78.9 \% accuracy) work best when combined with ML models rather than used alone. Results show that ML ensembles achieved the highest performance (95.78\% accuracy, ROC-AUC 0.96), while LLMs performed moderately in zero-shot (78.9\%) and slightly better in few-shot (72.6\%) settings. The proposed hybrid method enhanced the strength in uncertain situations, illustrating that ensemble ML is considered the best structured tabular prediction case, but it can be integrated with hybrid ML-LLM systems to provide a minor increase and open the way to more reliable clinical decision-support tools.
\end{abstract} 

\section{Introduction}
Cardiovascular disease (CVD) remains the leading cause of death worldwide and poses a serious preventive challenge to healthcare organizations \cite{george2023global}. Early and accurate risk stratification is therefore essential for reducing mortality. In contemporary clinical settings, CVD risk prediction is typically based on a limited number of structured variables, including demographics (age, sex), vital signs (blood pressure), biochemical markers (cholesterol, fasting glucose), and electrocardiogram (ECG)-based characteristics. However, these datasets tend to be highly imbalanced in favor of non-disease classes, which diminishes the sensitivity of predictive models to minority disease cases \cite{ahsan2022mlreview}.

Machine learning (ML) algorithms have demonstrated a strong capability to capture complex and non-linear patterns in clinical data without requiring strong parametric assumptions \cite{gorishniy2021tabular}. Ensemble models, including Random Forest, XGBoost, LightGBM, and CatBoost, have consistently outperformed traditional models such as logistic regression and support vector machines in case-by-case prediction of cardiovascular diseases (CVDs) \cite{tiwari2023ensemble, ali2021comparison}. These models exhibit favorable bias--variance trade-offs and robustness to noise when trained on tabular data. However, challenges related to model calibration and generalizability continue to limit their large-scale adoption in clinical practice \cite{frontai2025review}.

Although deep learning models are increasingly popular, empirical evidence suggests that shallow neural networks and Naive Bayes variants often perform poorly compared to gradient-boosted ensemble methods on small and highly imbalanced medical datasets \cite{mahajan2023ensemble, liu2025ehr}. Techniques such as Synthetic Minority Over-sampling Technique (SMOTE) and cost-sensitive learning can partially address class imbalance; however, they often fail to generalize effectively to external validation cohorts. Consequently, ensemble-based approaches have become the de facto standard for tabular medical classification tasks, including cardiovascular disease diagnostics.

Recently, large language models (LLMs) have emerged as zero-shot and few-shot predictors capable of interpreting structured and unstructured clinical data \cite{kablan2023stacked, cai2024ai}. When adapted to healthcare contexts, LLMs can extract semantic meaning from electronic health records and textual cues. However, their performance in highly imbalanced numeric datasets remains inconsistent and prompt-dependent, indicating that supervised ML frameworks currently maintain superior predictive reliability \cite{krittanawong2020ai, saito2015precisionrecall}. Current ensemble machine learning lacks robust reasoning on imbalanced or ambiguous clinical data, limiting predictive reliability. No prior framework integrates ML ensembles with LLM reasoning for structural heart disease prediction. We propose a novel ML–LLM fusion pipeline that merges statistical accuracy with interpretable, human-like decision support. This approach advances automated diagnosis by improving both predictive reliability and clinical transparency.

\section{Literature Review}
Many recent works have studied the ensemble and stacking protocols for the prediction of cardiovascular disease (CVD). Most authors have reported improvement over a single model baseline. However, the computational treatment of the stacking process has important drawbacks.  Mienye et al.\cite{mienye2020ensemble} developed a hybrid model using Random Forest and SVM classifiers to improve predictive performance; however, their work relied mostly on accuracy-based evaluation and did not check for calibration of probabilities or clinical reliability.  Purushottam et al. \cite{purushottam2016efficient} developed hybrid machine learning systems on UCI and Statlog datasets. However, their framework does not measure influence of the boosting strategy, imbalance sensitivity, and robustness across different evaluation settings. Recent research leveraged stacking with an optimized feature set to reach very high balanced accuracy  \cite{kumar2024stacked, mittal2025stacking, kumar2025stacked}. This shows that stacking is effective; however, the works primarily emphasize aggregate performance metrics and present little qualitative analysis of calibration quality, uncertainty, or decision reliability. Many other works have focused on various individual aspects in isolation. For instance, stability has been improved through feature selection \cite{praveen2024enhanced}. Similarly, interpretability has been enhanced through multi-criteria decision-making \cite{grace2025stacked}. The accuracy of models has been improved through hybrid deep learning ensembles \cite{alhussan2022hybrid}. Alternative ensemble models like collaborative clustering fusion \cite{araujo2023fusion}, stacking for heart attack diagnosis \cite{nguyen2024stacking}, and other early ensemble baselines \cite{sivasamy2024early} have also exhibited some noticeable gains, although generally with limited experimental designs and lacking common evaluation protocols. More recently, stacking methods focused on boosting precision did improve recall rates for the minority class \cite{sharma2025precisionstacking}. Similarly, explainable AI and meta-learning approaches improved the interpretability and calibration \cite{chen2025xai, wang2022stacking}. Just like that, Solaiman et al. \cite{solaiman2022x} confirmed that ensemble models are better than single classifiers achieving up to 96.25\% accuracy. Current work remains fragmented, optimizing an isolated objective (e.g., accuracy, recall, interpretability). Overall, there is no framework proposed that jointly optimizes for class imbalance, calibration reliability, and explainability. In contrast, the present work proposes a unified ensemble framework that jointly amalgamates imbalance-aware stacking, calibrated probability estimation, and interpretable reasoning as a means of going beyond accuracy-centric optimization for a more robust, trustworthy, and clinically relevant decision-support model.

\section{	METHODOLOGY}
\subsection{	Dataset Description and Preprocessing}\label{AA}
A publicly available heart disease dataset \cite{lapp_heart_disease_dataset_2019} dating from 1988, which was created by merging five independent repositories, namely Cleveland (303 records), Hungarian (294), Switzerland (123), Long Beach VA (200), and Stalog Heart (270), yielding a combined total of 1190 patient records, was used in this study. From the original 76 attributes, this study utilizes only 11 input features and one output feature (heart disease) shared across all sources (Table \ref{tab1}).
\begin{table}[htbp]
\caption{Dataset Attribution}
\begin{center}
\begin{tabular}{ l l }
\hline
\textbf{Attribute} & \textbf{Values / Range} \\
\hline
Age & Continuous \\
Sex & Male / Female \\
Chest Pain Type & TA, ATA, NAP, ASY \\
Resting BP & Continuous \\
Cholesterol & Continuous \\
Fasting BS & 1 = Yes, 0 = No \\

Resting ECG & Normal, ST, LVH \\

Max HR & 60–202 bpm \\

Exercise Angina & Y = Yes, N = No \\

Old Peak & Continuous \\

ST\_Slope & Up, Flat, Down \\

Heart Disease & 1 = Disease, 0 = No disease \\
\hline
\multicolumn{2}{l}{${\mathrm{}}$} 
\end{tabular}
\label{tab1}
\end{center}
\end{table}
For transparency and reproducibility, the five datasets were merged using 11 shared features with no missing values. Categorical variables were encoded, continuous features were normalized using MinMaxScaler, and missing values were appropriately imputed. The dataset was partitioned into stratified training, validation, and test sets (60/20/20), with SMOTE applied to address class imbalance using a fixed \texttt{random\_state = 42} to ensure reproducibility. All preprocessing steps were fitted exclusively on the training set to prevent information leakage.

\subsection{Machine Learning Models and Training Procedure}

In this study, we tested nine machine learning and deep learning models to classify heart disease. The models included CatBoost, Random Forest, XGBoost, LightGBM, Gradient Boosting, SVM, Logistic Regression, Multi-Layer Perceptron (MLP), and Naive Bayes. We used grid search to find the best hyperparameters for each model. Logistic Regression reached $84.03\%$ accuracy with $C=10$ and L2 regularization. Random Forest achieved $92.02\%$ accuracy with 300 estimators, a maximum depth of 7, a minimum split of 5, a minimum leaf size of 2, and the square root of the number of features as the maximum features. XGBoost had $90.76\%$ accuracy with a maximum depth of 7, a learning rate of 0.1, a subsample of 0.9, and $\gamma = 0.1$. LightGBM reached $90.34\%$ accuracy with a maximum depth of 5 and regularization parameters $\alpha = 0.1$ and $\lambda = 0.01$. CatBoost performed best, with $92.44\%$ accuracy using a depth of 5, 300 iterations, a learning rate of 0.1, and a temperature of 1. Gradient Boosting achieved $89.92\%$ accuracy with a depth of 5, a learning rate of 0.1, a subsample of 0.9, and a minimum of 2 samples. SVM reached $85.02\%$ accuracy with $C=10$, $\gamma=\text{scale}$, and the RBF kernel. Both MLP and Naive Bayes had $83.19\%$ accuracy. MLP used $(100, 50)$ ReLU layers, and Naive Bayes used variance smoothing of $1\times10^{-6}$. Overall, ensemble models such as CatBoost, Random Forest, XGBoost, LightGBM, and Gradient Boosting outperformed the simpler classifiers because they could capture more complex feature interactions and better balance bias and variance.

\subsection{Ensemble Machine Learning Voting Strategy}
\begin{figure}[htbp]
\centering
\includegraphics[width=1\linewidth]{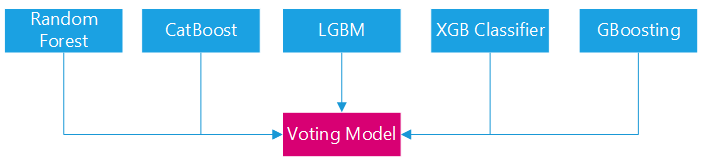}
\caption{ \textit{ML Voting Model}}
\label{fig:voting_model}
\end{figure}

First, the top five trained classifiers are gradient boosting, Random Forest, XGBoost, LightGBM, and CatBoost (Shown in Fig \ref{fig:voting_model}), were loaded using joblib. These pretrained models were combined into a Voting Classifier ensemble. The final prediction was decided by a majority vote among the five models using a soft and hard voting technique. This ensemble method enhances the robustness and predictive accuracy of heart disease classification when compared to standalone models.

\subsection{Zero-shot and few-shot LLMs}

This study evaluated ten large language models (LLMs) on zero-shot and few-shot classification tasks using accuracy and ROC-AUC as performance metrics. The top five models in zero-shot were Qwen3-Coder, Grok-Code-Fast, GLM-4.5-Air, LLaMA-4-Maverick, and Kimi-K2. For few-shot tasks, the leading models were LLaMA-4-Maverick, Kimi-K2, Mistral-24B, Qwen3-Coder, and GLM-4.5-Air. Each model was tested on a benchmark dataset without fine-tuning for zero-shot evaluation and with a few labeled examples for few-shot scenarios.

\subsection{LLM Ensemble Voting}\label{SCM}
For heart disease prediction, the top five LLMs were evaluated in zero-shot and few-shot settings. Predictions made by the model were loaded in CSV files, and the individual accuracy was calculated against the ground truth. The final class was determined in soft voting using the accuracy weighted and summed predictions thresholded at 0.5. In hard voting there was a combination of binary prediction and the outcome was decided by majority vote. The advantage of this ensemble strategy is that it enhances the reliability of a strategy through consensus and high-performing models without requiring further fine-tuning.

\subsection{Proposed Hybrid ML–LLM Fusion Framework}\label{AAA}
\begin{figure}[htbp]

        \centering
        \includegraphics[width=1\linewidth]{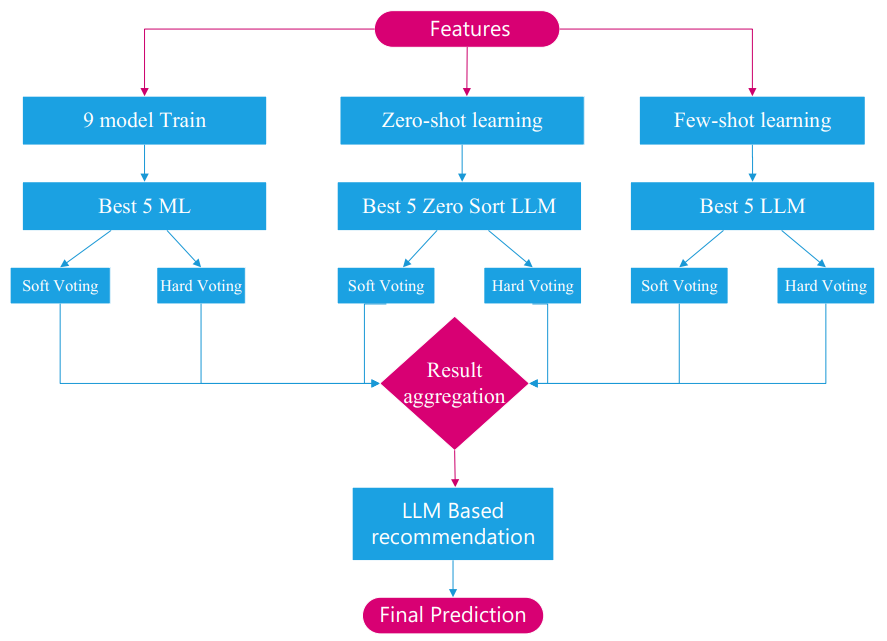}
        \caption{\textit{{\textit{Proposed Hybrid ML–LLM Fusion Framework}}}}
        \label{fig:placeholder}
    \end{figure}

{The prediction system (Shown in Fig \ref{fig:placeholder})  used three voting ensembles: top five ML models (soft/hard accuracy: 0.95/0.93), top five LLM zero-shot (0.776/0.776) and top five LLM few-shot (0.726/0.726).For a sample patient, the ML models predicted heart disease while the LLM models predicted no disease. Weighting predictions by model accuracy produced a risk score of 0.77, indicating a high probability of heart disease. This aggregation demonstrates the influence of high-accuracy ML models in ensemble decision-making. The system provides both a quantitative risk score and advisory guidance for research purposes.}

\section{	Result}
\subsection{Performance of Individual Machine Learning Models}
\begin{figure}
    \centering
    \begin{minipage}[b]{0.80\linewidth}
        \centering
        \includegraphics[width=\linewidth]{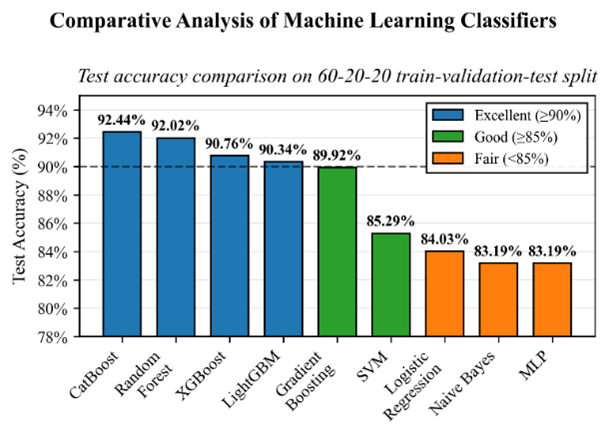}
        %\vspace{0pt} % no space between image and next
    \end{minipage}
   
    \caption{\textit{ Model test accuracy comparison for all machine learning models}}
    \label{fig:accuracy-roc1}
    
\end{figure}

\begin{figure}
    \centering
    \begin{minipage}[b]{0.80\linewidth}
        \centering
        \includegraphics[width=\linewidth]{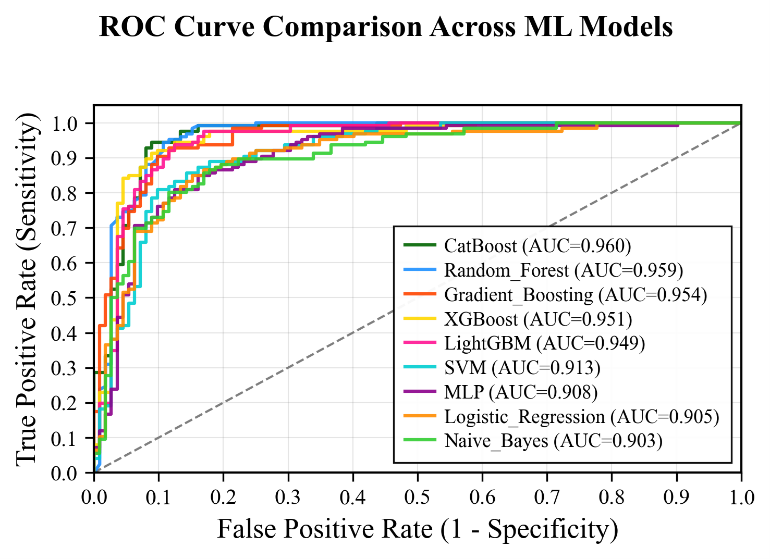}
        %\vspace{0pt} % no space between image and next
    \end{minipage}
   
    \caption{\textit{ ROC curves comparison for all machine learning models}}
    \label{fig:accuracy-roc2}

\end{figure}

The initial step involved an analysis of the performance metrics associated with nine classical machine learning classifiers applied to the dataset.  Each model was trained on a stratified 60/20/20 split. In terms of pre-processing, the training set underwent Min-Max scaling and SMOTE application. \textbf{Table} \ref{ITH} shows the accuracy measured on test set. ML model test accuracy and ROC curves comparison are shown in Fig \ref{fig:accuracy-roc1} and Fig \ref{fig:accuracy-roc2}, respectively.

\begin{table}[!ht]
\caption{\textit{Performance of Individual Machine Learning Models}}

\label{ITH}
\centering
\begin{tabular}{ l c }

\hline
\textbf{Model} & \textbf{Accuracy (\%)} \\

\hline
\textbf{CatBoost)} & \textbf{92.44} \\
Random Forest & 92.02 \\
XGBoost & 90.76 \\
LightGBM & 90.34 \\
Gradient Boosting & 89.92 \\
SVM & 85.02 \\
Logistic Regression & 84.03 \\
MLP & 83.19 \\
Naïve Bayes & 83.19 \\
\hline

\end{tabular}

\vspace{4pt} % add space between table and caption

\end{table} Ensemble models based on trees outperformed classical baselines at all levels by maintaining n size, with CatBoost achieving the highest accuracy (92.44\%). The results with Naive Bayes and MLP were considerably low because training deeper architectures on small (tabular) datasets is a big challenge.

\subsection{Ensemble Machine Learning Results}
\begin{figure}[!t]
    \centering
    \begin{minipage}[b]{0.48\linewidth}
        \centering
        \includegraphics[width=\linewidth]{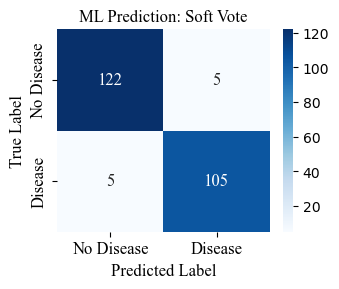}
    \end{minipage}
    \hfill
    \begin{minipage}[b]{0.48\linewidth}
        \centering
        \includegraphics[width=\linewidth]{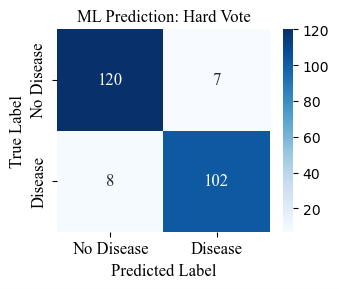}
    \end{minipage}
    \caption{\textit{Confusion matrices of ensemble predictions: soft voting (left) and hard voting (right)}}
    \label{fig:ensemble-confusion-wide}
\end{figure}

To boost the predictability stability further, we also created ensembles on the five best-classified models are Random Forest, XGBoost, CatBoost, LightGBM, and Gradient Boosting. We implemented soft voting and weighted soft voting with talents proportional to validating AUC. \textbf{Table} \ref{Ensemble Model Results} presents the ensemble performance. Confusion matrices of ensemble performance are shown in Fig \ref{fig:ensemble-confusion-wide}.

\begin{table}[!t]

\begin{center}
\caption{\textit{Ensemble Model Results}}
\begin{tabular}{l c c }
\hline
\textbf{Ensemble Strategy} & \textbf{Accuracy (\%)} & \textbf{AUC} \\
\hline
\textbf{Soft Voting (Weighted))} & \textbf{95.23}  & 0.96\\

Hard Voting & 93.78 & 0.98 \\
\hline

\end{tabular}
\label{Ensemble Model Results}
\end{center}
\vspace{-6mm}
\end{table}

The stacking ensemble produced an overall best performance with an accuracy of 95.23\% and an AUC of 0.96. This implies that the ensemble and mix of multiple strong learners enhance robustness and generalization over any model.
\subsection{Large Language Model Predictions}
Simultaneously, we also tested the cutting-edge LLMs using Open Router APIs. Zero- and few-shot prompting were used to make predictions. The standardized tabular schema was used to prompt the models to produce binary disease labels.\textbf{Table} IV represents the performance of LLMs with zero-shot and few-shot prompting. 
\begin{table}[htbp]
\label{tablep}
\vspace{2pt}
\caption{\textit{Performance of LLMs with Zero-Shot and Few-Shot Prompting}}
\begin{center}

% Compact layout settings
\setlength{\tabcolsep}{4pt} % reduce horizontal cell padding
\renewcommand{\arraystretch}{0.85} % reduce vertical row spacing

\begin{tabular}{ l c c }
\hline
\textbf{Model Name} & \textbf{Accuracy (\%)} & \textbf{Accuracy (\%)} \\
\hline
\textbf{Qwen 3 Coder} & \textbf{81.34} & 72.94 \\
X AI Grok Code Fast 1 & 81.18 & 71.51 \\
Z AI GLM 4.5 Air & 80.08 & 72.94 \\
Meta Llama 4 Maverick & 79.24 & 76.20 \\
Moonshot AI Kimi K2 & 78.82 & 75.29 \\
OpenAI GPT 5 & 77.39 & 52.86 \\
Mistral Small 3.2 24B Instruct & 75.21 & 73.50 \\
NVIDIA Nemotron Nano 9B V2 & 64.03 & 62.10 \\
\textbf{Deepseek V3.1} & 79.16 & \textbf{80.73} \\
Gemini 2.5 Flash & 77.12 & 77.06 \\

\hline

\end{tabular}

\end{center}
\vspace{-5mm}
\end{table}
\subsection{	LLM Voting Result }
Soft voting was used to further test ensemble strategies with large language model (LLM) features under two settings. The few-shot soft and hard voting models demonstrated accuracies of 72.6\% and 72.2\%, accompanied by ROC-AUC scores of 0.727 and 0.729, respectively. In contrast, the zero-shot soft and hard voting models demonstrated superior performance, achieving accuracies of 78.9\% and 77.6\%, respectively, along with ROC-AUC values of 0.804 and 0.782. These findings suggest that a zero-shot soft voting has a more valid classification performance than few-shot, implying that zero-shot is more effective in LLM-based predictions, which is clearly depicted in Table \ref{tab:llm_voting_results}  and Fig. \ref{Confusion matrices of LLM voting results under (a) Zero Shot(Soft vote), (b) Zero Shot (Hard vote), (c) Few Shot (Soft vote) and (d) Few Shot (Hard vote) }

\begin{table}[!ht]
\vspace{1pt}\caption{\textit{LLM Model Voting Results}}
\label{tab:llm_voting_results}
\centering
\begin{tabular}{lcc}
\hline
\textbf{Method} & \textbf{Accuracy} & \textbf{AUC} \\
\hline
\textbf{Zero soft voting} & \textbf{0.789} & \textbf{0.804} \\
Zero hard voting & 0.776 & 0.782 \\
Few soft voting  & 0.726 & 0.727 \\
Few hard voting  & 0.722 & 0.729 \\
\hline
\end{tabular}
\end{table}

\begin{figure}[!ht]
    \centering
    \begin{minipage}{1.00\linewidth}
        \centering
        \includegraphics[width=\linewidth]{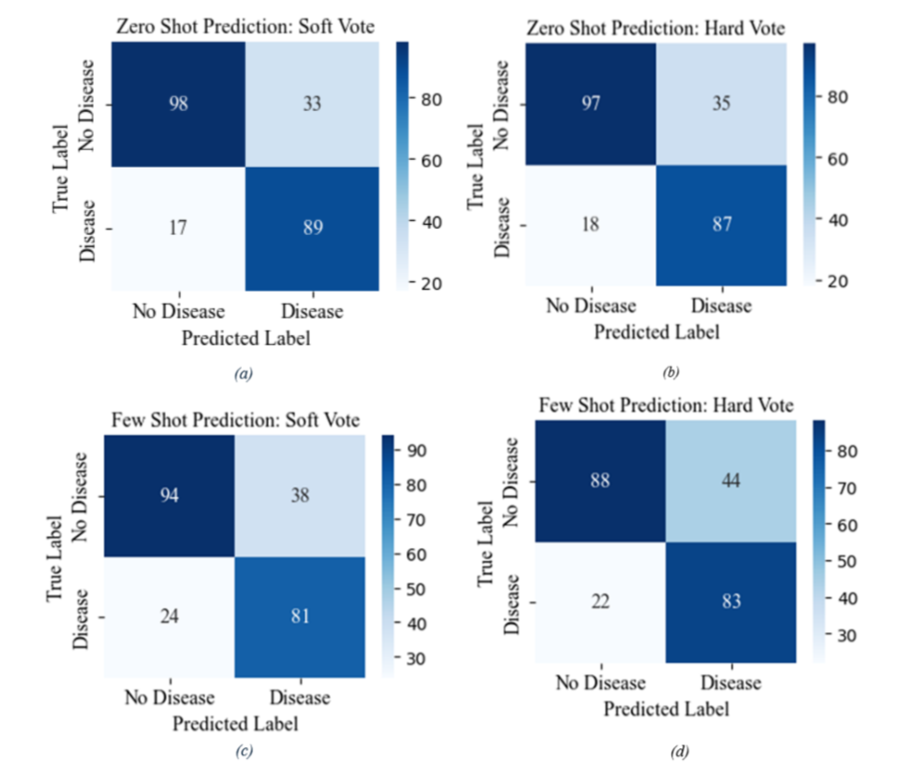}
        
    \end{minipage}
   
    \caption{\textit{{Confusion matrices of LLM voting results under (a) Zero Shot(Soft vote), (b) Zero Shot (Hard vote), (c) Few Shot (Soft vote) and (d) Few Shot (Hard vote)}}}
    \label{Confusion matrices of LLM voting results under (a) Zero Shot(Soft vote), (b) Zero Shot (Hard vote), (c) Few Shot (Soft vote) and (d) Few Shot (Hard vote) }
\end{figure}

\subsection{ML–LLM Fusion Pipeline Results and Comparison}
It was tested using a hybrid fusion framework that fuses probabilistic outputs of the machine learning ensemble with the prediction of GPT-4.1 (few-shot), and Gemini 2.5 Flash acts as a meta-reasoning layer to combine both these outputs to resolve ambiguity in the caertainty in the statistical classifiers.

\vspace{-4mm}
\begin{table}[htbp]
\label{tab6}
\vspace{1pt}\caption{\textit{Comparative Performance of Ensemble-Based CVD Prediction Methods}}
\begin{center}
\begin{tabular}{ l l c }
\hline
\textbf{Ref.} & \textbf{Model / Approach} & \textbf{Accuracy (\%)} \\
\hline
{[16]} & Feature-Selected Stacking Ensemble & 82.56 \\
{[18]} & AdaBoost Decision Fusion (ABDF) & 83.0 \\
{[14]} & Data Mining Model (10-fold CV) & 86.3 \\
{[15]} & Firefly + Stacking Ensemble & 86.79 \\
{[25]} & Stacking Ensemble \& Voting Ensemble & 91.0 \\
{[17]} & VAE + DNN Stacked Ensemble (HDPM) & 92.3 \\
{[13]} & Weighted Aging CART Ensemble & 93.0 \\
\textbf{Our Work} & \textbf{Fusion (ML + LLM)} & \textbf{96.62} \\
\hline
\end{tabular}
\end{center}
\end{table}

\vspace{-6mm}
\begin{figure}[!ht]
    \centering
    \begin{minipage}[b]{0.60\linewidth}
        \centering
        \includegraphics[width=\linewidth]{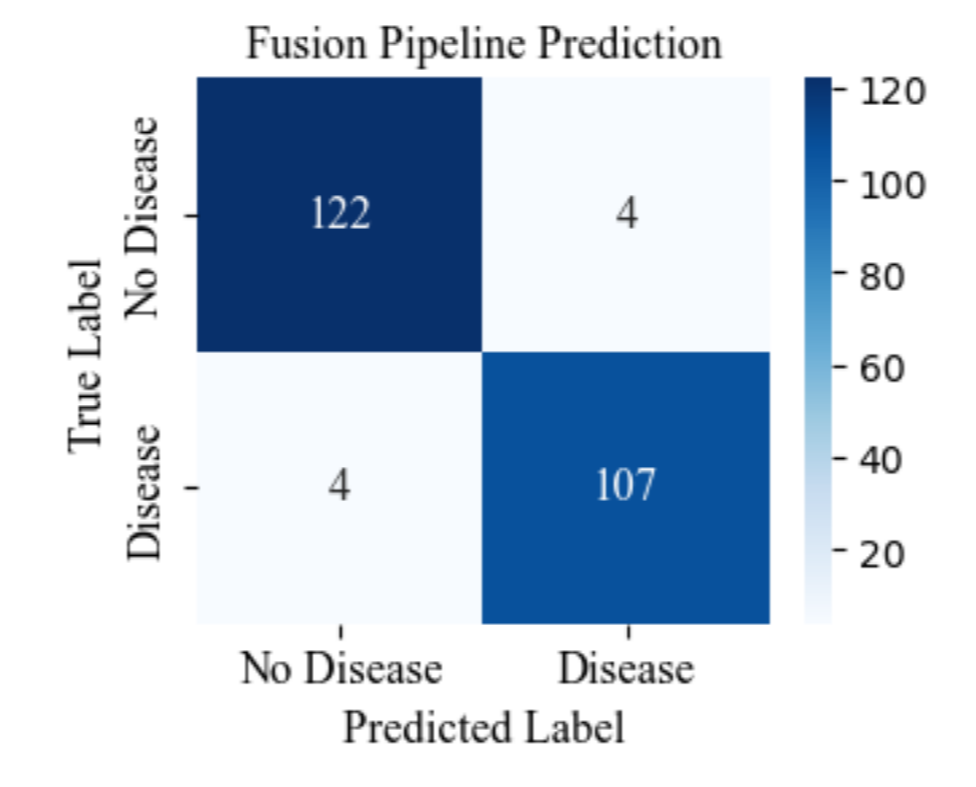}
        %\vspace{0pt} % no space between image and next
    \end{minipage}
   
    \caption{\textit{ Confusion matrices of fusion Pipeline }}
    \label{fig:accuracy-roc3}
    
\end{figure}  

Table VI is a comparative analysis, which reveals that the proposed hybrid fusion framework has a low but steady accuracy (96.62\%) and ROC-AUC (0.97) compared to the past ensemble techniques. This gain is primarily attributed to the integration of LLM-based contextual reasoning, which refines predictions in cases where statistical classifiers are uncertain. LLMs are employed exclusively as auxiliary decision-support modules, complementing rather than replacing traditional machine learning ensembles.

\subsection{Error Analysis}
\begin{figure}[!ht]
    \centering
    \begin{minipage}{0.80\linewidth}
        \centering
        \includegraphics[width=\linewidth]{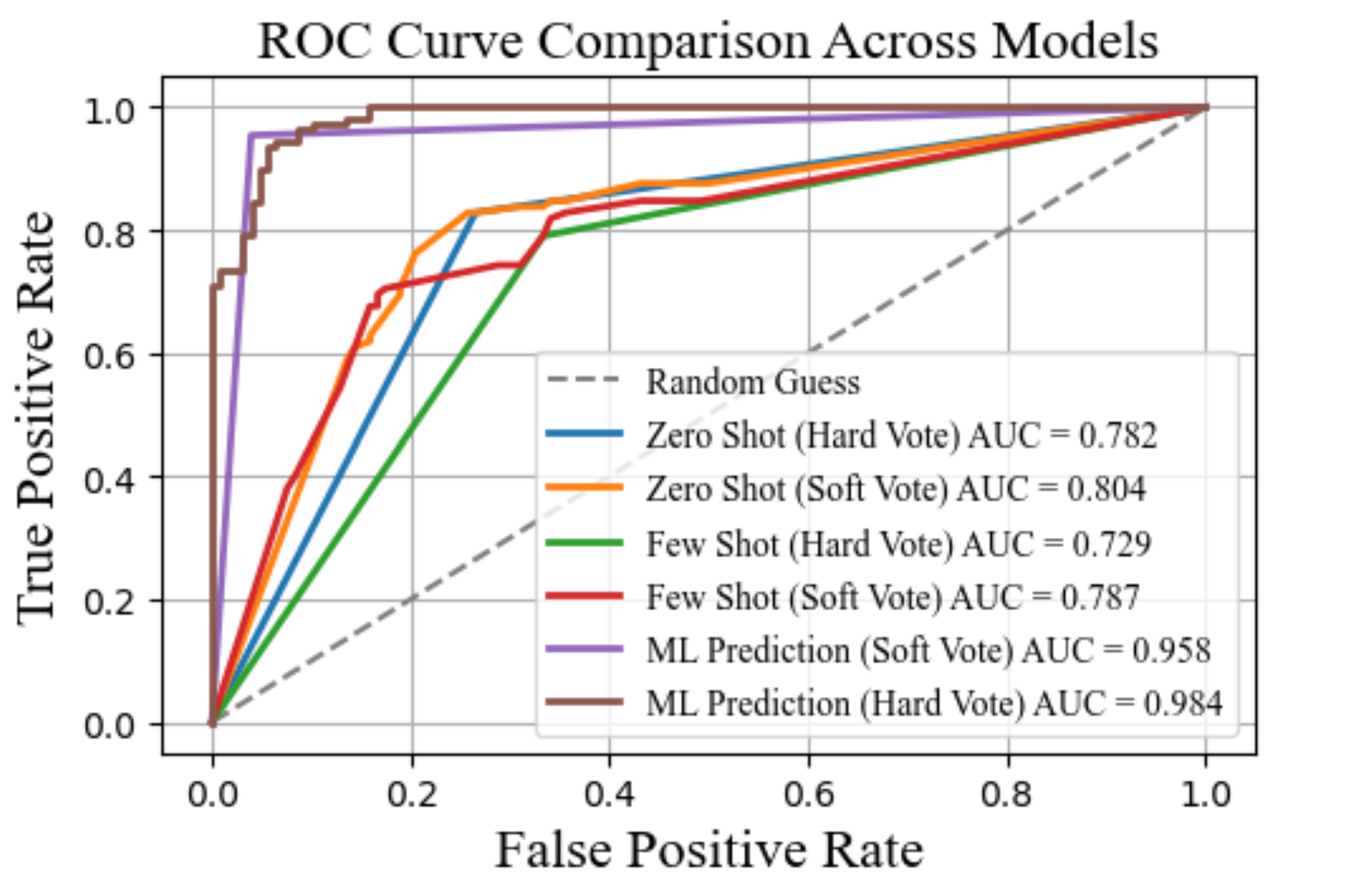}
        
    \end{minipage}
   
    \caption{\textit{ROC Curves of Comparison Across Model.}}
    \label{fig:roc_curves}
\end{figure}

Error analysis reveals that ML ensembles achieve balanced sensitivity and specificity, effectively identifying most cases of the rare disease class. However, the misclassification rates of this class are lower with LLMs. Due to the elimination of errors and increased performance, especially on borderline and rare cases Fig. \ref{fig:roc_curves}, getting ML ensembles together with LLMs further improves the performance of these models.

\subsection{Limitations and Future Work}
The main limitation of the study is the small harmonized cohort (1,190 records) taken to Cleveland and Hungary data which limits the extrapolation of the results. The analysis was based on accuracy, ROC-AUC, and confusion matrices, without calibration and a cost-sensitive analysis. Advanced machine learning algorithms and LLMs were considered models and were mostly tested using zero- and few-shot prompting, with variation in repeated queries also noted with LLMs. Future studies will be shofocused on scaling to bigger and more multifarious datasets, introducing solid evaluation procedures, and enhancing hybrid ML-LLM methods, specifically with regard to clinical incorporation towards safety and equity.

\section{Discussion}
The comparison of traditional machine learning ensembles and large language models (LLMs) to predict heart disease with the help of structured clinical data was conducted. The tree-based ensemble models were able to achieve high accuracy of more than 90 percent and high ROC-AUC of more than 0.96. The five most successful models as a soft voting ensemble performed best (95.78\% accuracy, 0.96 ROC-AUC) and were stronger, more generalizable, and applicable in a clinical practice. The results highlight the long-term usefulness of tree based ensembles to the systematic forecasting of cardiovascular risks.

There are inherent weaknesses in the immediate application of LLLM to raw tabular data, such as the large variance between zero-shot and few-shot results on repeated queries. Even though there are minor gains in case of voting-based aggregation, their results are worse than those of ensemble machine learning models. Remarkably, the accuracy of LLMs is much higher when the tabular data is converted to the textual forms, which demonstrates that the input form is a crucial factor to the successful model recognition. These results indicate that more efforts are required to optimize LLMs to structured data domains.

\section{	Conclusion}
	This paper provides a comparative study of ensemble machine learning models and large language models (LLMs) in the field of heart disease prediction using more than 1,190 clinical cases. The tree-based ensemble classifiers, especially a soft voting ensemble of the five best models, were shown to be the best in cases with structured tabular data (95.78\%, 0.96 ROC-AUC). By comparison, LLMs were not very effective and predictable when used on raw tabular data. Nevertheless, there was the prospect of LLM in reasoning and interpretability, and the better performance when presented with semantically structured textual information. The findings outline the existing strengths and weaknesses of the two methods in cardiovascular risk prediction and emphasize the applicability of ensemble ML models in clinical work and interpretability of LLMs.

\bibliographystyle{IEEEtran}
\bibliography{references}

\clearpage
\onecolumn
\section*{\textbf {Author Contribution Statements}}

\noindent \textbf{Md. Tahsin Amin$^{1st}$ Author}

Md. Tahsin Amin served as the primary contributor to this study and led the Machine Learning (ML) component. He was responsible for the conceptualization and development of the proposed methodology, including designing and structuring the overall ML framework. He contributed significantly to the zero-shot learning component, formulated the experimental design, and conducted comprehensive error analysis to evaluate model performance and identify areas for improvement. In addition, he performed extensive manuscript review and multiple rounds of revision, corrected technical and grammatical issues, enhanced clarity and coherence, and conducted plagiarism screening to ensure originality and compliance with publication standards. 

\vspace{0.5cm}

\noindent \textbf{Tanim Ahmmod$^{1st}$ Author (Co--First)}

Tanim Ahmmod developed the final fusion framework and conducted the fusion-based experiments to achieve optimal performance. He implemented the zero-shot and few-shot approaches using Large Language Models (LLMs), contributed to the initial section of the methodology, designed all flowcharts presented in the manuscript, analyzed experimental results, and reviewed the manuscript to eliminate technical and structural inconsistencies.

\vspace{0.5cm}
\noindent \textbf {Md. Tahsin Amin $^{\dagger}$  and Tanim Ahmmod $^{\dagger}$} contributed equally to this work and share first authorship.

\vspace{0.5cm}

\noindent \textbf{Zannatul Ferdus$^{2nd}$ Author}

\vspace{0.5cm}

Zannatul Ferdus prepared the complete Latex manuscript and authored the Abstract, Introduction, Literature Review, Results, Discussion, and Future Work sections. She conducted thorough revisions to improve organization, clarity, and overall coherence of the manuscript.

\vspace{0.5cm}

\noindent \textbf{Talukder Naemul Hasan Naem$^{3rd}$ Author}

Talukder Naemul Hasan Naem assisted in debugging the zero-shot LLM implementation to ensure proper execution and reliable result generation. He contributed to editing and proofreading the entire manuscript, improved logical flow, refined the Abstract and Introduction sections, and ensured that the final document complied with submission guidelines.

\vspace{0.5cm}

\noindent \textbf{Ehsanul Ferdous$^{4th}$ Author}

Ehsanul Ferdous wrote the Methodology section and contributed to organizing the manuscript to ensure a clear and logical presentation of the research process. He also proofread and edited the text to improve readability and conciseness.

\vspace{0.5cm}

\noindent \textbf{Arpita Bhattacharjee$^{4th}$ co-Author}

Arpita Bhattacharjee conducted the few-shot Large Language Model (LLM) experiments and contributed to the evaluation and comparative analysis of different LLM models.

\vspace{0.5cm}

\noindent \textbf{Ehsanul Ferdous$^{*}$ and Arpita Bhattacharjee$^{*}$ contributed equally to this work and share fourth authorship.}

\vspace{0.5cm}

\noindent \textbf{Ishmam Ahmed Solaiman$^{5th}$ and Nahiyan Bin Noor$^{6th}$ Authors (Supervisors)}

They served as research supervisors and provided overall academic guidance and continuous monitoring throughout the research process. Their contributions included refining the research problem, strengthening the methodological design, offering critical feedback on the ML framework and zero-shot learning components, reviewing experimental findings, and providing constructive recommendations to enhance analysis, interpretation, and scientific rigor. Their supervision significantly improved the overall quality and presentation of the manuscript.

\end{document}